\documentclass[11pt,letterpaper]{article}
\usepackage{emnlp2017}
\usepackage{times}
\usepackage{latexsym}
\usepackage{tikz}

\usepackage{microtype} 

\emnlpfinalcopy

\def\emnlppaperid{***}

\usepackage{longtable}
\newcommand{\numRuns}[0]{50.000}

\usepackage{graphicx}
\usepackage{caption}
\usepackage{subcaption}
\usepackage{multirow}
\usepackage{placeins}
\usepackage{amsmath}
\usepackage{amssymb}
\usepackage[parfill]{parskip}

\usepackage{array}
\newcolumntype{P}[1]{>{\centering\arraybackslash}p{#1}}
  
\newcommand{\refsec}[1]{\hyperref[#1]{section \ref*{#1}}}
\newcommand{\refsecUpper}[1]{\hyperref[#1]{Section \ref*{#1}}}
\newcommand{\reftable}[1]{\hyperref[#1]{Table \ref*{#1}}}
\newcommand{\reffigure}[1]{\hyperref[#1]{Figure \ref*{#1}}}

\title{Reporting Score Distributions Makes a Difference: Performance Study of LSTM-networks for Sequence Tagging}

\author{Nils Reimers \and Iryna Gurevych \\
Ubiquitous Knowledge Processing Lab (UKP) and Research Training Group AIPHES\\
Department of Computer Science, Technische Universit\"at Darmstadt\\
Ubiquitous Knowledge Processing Lab (UKP-DIPF) \\ 
German Institute for Educational Research \\
\url{www.ukp.tu-darmstadt.de}}

\date{}

\newcommand\blfootnote[1]{%
  \begingroup
  \renewcommand\thefootnote{}\footnote{#1}%
  \addtocounter{footnote}{-1}%
  \endgroup
}

\begin{document}

\maketitle

\begin{abstract}
In this paper\blfootnote{This paper is published in the proceedings of EMNLP 2017.} we show that reporting a single performance score is insufficient to compare non-deterministic approaches. We demonstrate for common sequence tagging tasks that the seed value for the random number generator can result in \textit{statistically significant} ($p < 10^{-4}$) differences for state-of-the-art systems. For two recent systems for NER, we observe an absolute difference of one percentage point $F_1$-score depending on the selected seed value, making these systems perceived either as \textit{state-of-the-art} or \textit{mediocre}. Instead of publishing and reporting single performance scores, we propose to compare score distributions based on multiple executions. 
Based on the evaluation of \numRuns{} LSTM-networks for five sequence tagging tasks, we present network architectures that produce both superior performance as well as are more stable with respect to the remaining hyperparameters. The full experimental results are published in \cite{Reimers_2017_Hyperparameter}.\footnote{\url{https://arxiv.org/abs/1707.06799}} The implementation of our network is publicly available.\footnote{\url{https://github.com/UKPLab/emnlp2017-bilstm-cnn-crf}} 
\end{abstract}

\section{Introduction} \label{intro}
Large efforts are spent in our community on developing new \textit{state-of-the-art} approaches. To document that those approaches are better, they are applied to unseen data and the obtained performance score is compared to previous approaches. In order to make results comparable,  a provided split between train, development and test data is often used, for example from a former shared task.

In recent years, deep neural networks were shown to achieve state-of-the-art performance for a wide range of NLP tasks, including many sequence tagging tasks \cite{Ma2016}, dependency parsing \cite{Syntaxnet}, and machine translation \cite{GoogleGMT}. The training process for neural networks is highly non-deterministic as it usually depends on a random weight initialization, a random shuffling of the training data for each epoch, and repeatedly applying random dropout masks. The error function of a neural network is a highly non-convex function of the parameters with the potential for many distinct local minima \cite{LeCun1998,Erhan2010}. Depending on the seed value for the pseudo-random number generator, the network will converge to a different local minimum.

Our experiments show that these different local minima have vastly different characteristics on unseen data. For the recent NER system by \newcite{Ma2016} we observed that, depending on the random seed value, the performance on unseen data varies between $89.99\%$ and $91.00\%$  $F_1$-score. The difference between the best and worst performance is statistically significant ($p < 10^{-4}$) using a randomization test\footnote{1 Million iterations. p-value adapted using the Bonferroni correction to take the 86 tested seed values into account.}. In conclusion, whether this newly developed approach is perceived as \textit{state-of-the-art} or as \textit{mediocre}, largely depends on which random seed value is selected. This issue is not limited to this specific approach, but potentially applies to all approaches with non-deterministic training processes.

This large dependence on the random seed value creates several challenges when evaluating new approaches:
\begin{itemize}
 \item Observing a (statistically significant) improvement through a new non-deterministic approach  might not be the result of a superior approach, but the result of having a more favorable sequence of random numbers.
 \item Promising approaches might be rejected too early, as they fail to deliver an outperformance simply due to a less favorable sequence of random numbers.
 \item Reproducing results is difficult.   
\end{itemize}

To study the impact of the random seed value on the performance we will focus on five linguistic sequence tagging tasks:  POS-tagging, Chunking, Named Entity Recognition, Entity Recognition\footnote{Entity Recognition labels all tokens that refer to an entity in a sentence, also generic phrases like \textit{U.S. president}.}, and Event Detection. Further we will focus on Long-Short-Term-Memory (LSTM) Networks \cite{Hochreiter1997}, as those demonstrated state-of-the-art performance for a wide variety of sequence tagging tasks \cite{Ma2016,Lample2016,Sogaard2016}. 

Fixing the random seed value would solve the issue with the reproducibility, however, there is no justification for choosing one seed value over another seed value. Hence, instead of reporting and comparing a single performance, we show that comparing score distributions can lead to new insights into the functioning of algorithms.  

Our main contributions are:
\begin{enumerate}
 \item Showing the implications of non-deterministic approaches on the evaluation of approaches and the requirement to compare score distributions instead of single performance scores. 
 \item Comparison of two recent, state-of-the-art systems for NER and showing that reporting a single performance score can be misleading.
 \item In-depth analysis of different LSTM-architectures for five sequence tagging tasks with respect to: superior performance, stability of results, and importance of tuning parameters. 
\end{enumerate}

\section{Background}
Validating and reproducing results is an important activity in science to manifest the correctness of previous conclusions and to gain new insights into the presented approaches. \newcite{Fokkens2013} show that reproducing results is not always straight-forward, as factors like preprocessing (e.g.\ tokenization), experimental setup (e.g.\ splitting data), the version of components, the exact implementation of features, and the treatment of ties can have a major impact on the achieved performance and sometimes on the drawn conclusions. 

For approaches with non-deterministic training procedures, like neural networks, reproducing exact results becomes even more difficult, as randomness can play a major role in the outcome of experiments. The error function of a neural network is a highly non-convex function of the parameters with the potential for many distinct local minima \cite{LeCun1998,Erhan2010}. The sequence of random numbers plays a major role to which minima the network converges during the training process. However, not all minima generalize equally well to unseen data. \newcite{Erhan2010} showed for the MNIST handwritten digit recognition task that different random seeds result in largely varying performances. They noted further that with increasing depth of the neural network, the probability of finding poor local minima increases.

\begin{figure}[htp]
\centering
\begin{tikzpicture}
\draw[step=1cm,gray,very thin,dotted] (-2,-0.5) grid (3.5,3);
\draw (0,0) parabola (2.0,3);
\draw (0,0) parabola (-2,3);
\draw (2.5,0) parabola (2,3);
\draw (2.5,0) parabola (3.0,3);
\draw[dashed] (0,-0.5) -- (0,3);
\draw[dashed] (2.5,-0.5) -- (2.5,3);
\draw[red,dashed] (0.5,0) parabola (2.5,3);
\draw[red,dashed] (0.5,0) parabola (-1.5,3);
\draw[red,dashed] (3.0,0) parabola (2.5,3);
\draw[red,dashed] (3.0,0) parabola (3.5,3);
\filldraw[black] (0,-0.5)  node[anchor=north] {Flat Minimum};
\filldraw[black] (2.5,-0.5)  node[anchor=north] {Sharp Minimum};
\filldraw[black] (-3.0,3) circle(0pt) node[anchor=south west] {Train/Dev Error};
\filldraw[red] (2.5,3) circle(0pt) node[anchor=south] {Test Error};
\filldraw[black] (-2,2) circle(0pt) node[anchor=east] {$f(x)$};
\end{tikzpicture}
\caption{A conceptual sketch of flat and sharp minima from \newcite{Keskar2016}. The Y-axis indicates values of the error function and the X-axis the weight-space. \label{fig:sharp_flat_minima}}
\end{figure}
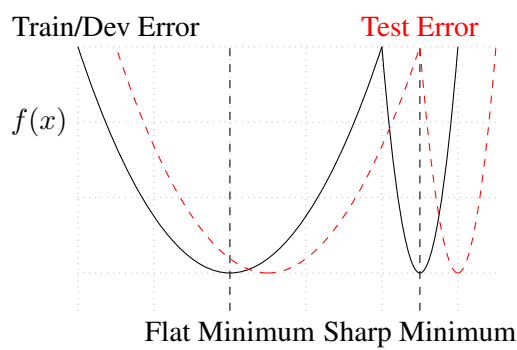

As (informally) defined by \newcite{Hochreiter1997_Flat_Minima}, a minimum can be flat, where the error function remains approximately constant for a large connected region in weight-space, or it can be sharp, where the error function increases rapidly in a small neighborhood of the minimum. A conceptual sketch is given in \reffigure{fig:sharp_flat_minima}. The error functions for training and testing are typically not perfectly synced, i.e.\ the local minima on the train or development set are not the local minima for the held-out test set. A sharp minimum usually depicts poorer generalization capabilities, as a slight variation results in a rapid increase of the error function. On the other hand, flat minima generalize better on new data \cite{Keskar2016}. Keskar et al.\ observe for the MNIST, TIMIT, and CIFAR dataset, that the generalization gap is not due to \textit{over-fitting} or \textit{over-training}, but due to different generalization capabilities of the local minima the networks converge to.

A priori it is unknown to which type of local minimum a neural network will converge. Some methods like the weight initialization \cite{Erhan2010,Glorot2010} or small-batch training \cite{Keskar2016} help to avoid bad (e.g.\ sharp) minima. Nonetheless, the non-deterministic behavior of approaches must be considered when they are evaluated.

\section{Impact of Randomness in the Evaluation of Neural Networks}\label{sec:evaluation_networks}

Two recent, state-of-the-art systems for NER are proposed by \newcite{Ma2016}\footnote{\url{https://github.com/XuezheMax/LasagneNLP}} and by \newcite{Lample2016}\footnote{\url{https://github.com/glample/tagger}}. Lample et al. report an $F_1$-score of $90.94\%$ and Ma and Hovy report an $F_1$-score of $91.21\%$. Ma and Hovy draw the conclusion that their system achieves a significant improvement over the system by Lample et al.

\begin{figure}[h]
\centering
  \includegraphics[width=.35\textwidth]{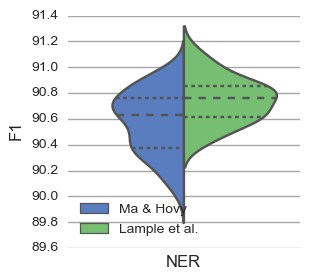}
\caption{Distribution of scores for re-running the system by Ma and Hovy (left) and Lample et al.\ (right) multiple times with different seed values. Dashed lines indicate quartiles. }
\label{fig:ma_lample_violin_plot}
\end{figure}

We re-ran both implementations multiple times, each time only changing the seed value of the random number generator. We ran the Ma and Hovy system 86 times and the Lample et al. system, due to its high computational requirement, for 41 times. The score distribution is depicted as a violin plot in \reffigure{fig:ma_lample_violin_plot}. Using a Kolmogorov-Smirnov significance test \cite{Massey1951}, we observe a statistically significant difference between these two distributions ($p < 0.01$). The plot reveals that the quartiles for the Lample et al.\ system are above those of the Ma and Hovy system. Further it reveals a smaller standard deviation $\sigma$ of the $F_1$-scores for the Lample et al.\ system. Using a Brown-Forsythe test, the standard deviations are different with $p < 0.05$.  \reftable{table:ma_vs_lample} shows the minimum, the maximum, and the median performance for the test performances.

\begin{table*}[t]
\centering
\begin{tabular}{|c|c||c|c|c|c|c|}
\hline
\textbf{System} & \textbf{Reported $F_1$} & \textbf{\# Seed values} & \textbf{Min. $F_1$} & \textbf{Median $F_1$} & \textbf{Max. $F_1$} & $\sigma$    \\ \hline
Ma and Hovy & $91.21\%$ & 86 & $89.99\%$ & $90.64\%$ & $91.00\%$ & 0.00241 \\ \hline
Lample et al. & $90.94\%$ & 41 & $90.19\%$ & $90.81\%$ & $91.14\%$ & 0.00176  \\ \hline
\end{tabular}
\caption{The system by \newcite{Ma2016} and \newcite{Lample2016} were run multiple times with different seed values.}
\label{table:ma_vs_lample}
\end{table*}

\begin{table*}[t]
\centering
\begin{tabular}{|c|c|c|c|c|c|}
\hline
\textbf{Task} & \textbf{Dataset} & \textbf{\# Configs} & \textbf{Median Difference} & \textbf{95th percentile} & \textbf{Max. Difference}   \\ \hline
POS & Penn Treebank & 269 & 0.17\% & 0.78\% & 1.55\% \\ \hline
Chunking & CoNLL 2000 & 385 & 0.17\% & 0.50\% & 0.81\% \\ \hline
NER & CoNLL 2003 & 406 & 0.38\% & 1.08\% & 2.59\% \\ \hline
Entities & ACE 2005 & 405 & 0.72\% & 2.10\% & 8.23\% \\ \hline
Events & TempEval 3  & 365 & 0.43\% & 1.23\% & 1.73\% \\ \hline
\end{tabular}
\caption{The table depicts the median, the 95th percentile and the maximum difference between networks with the same hyperparameters but different random seed values.}
\label{table:random_init}
\end{table*}

Based on this observation, we draw the conclusion that the system by Lample et al.\ outperforms the system by Ma and Hovy, as their implementation achieves a higher score distribution and shows a lower standard deviation. 

In a usual setup, approaches would be compared on a development set and the run with the highest development score would be used for unseen data, i.e.\ be used to report the test performance. For the Lample et al.\ system we observe a Spearman's rank correlation between the development and the test score of $\rho = 0.229$. This indicates a weak correlation and that the performance on the development set is not a reliable indicator. Using the run with the best development score ($94.44\%$) would yield a test performance of mere $90.31\%$. Using the second best run on development set ($94.28\%$), would yield state-of-the-art performance with $91.00\%$. This difference is statistically significant ($p < 0.002$). In conclusion, a development set will not necessarily solve the issue with bad local minima.

The main difference between these two approaches is in the generation of character-based representations: Ma and Hovy uses a Convolutional Neural Network (CNN) \cite{LeCun1989}, while Lample et al.\ uses an LSTM-network. As our experiments in \refsec{sec:char_representation} show, both approaches perform comparably if all other parameters were kept the same. Further, we could only observe a statistically significant improvement for the tasks POS, Chunking and Event Detection. For NER and Entity Recognition, the difference was statistically not significant given the number of tested hyperparameters.

In the next step, we evaluated the impact of the random seed value for the five  sequence tagging tasks described in \refsec{sec:experiment_setup}. We sampled randomly 1830 different configurations, for example different numbers of recurrent units, and ran the network twice, each time with a different seed value. The results are depicted in \reftable{table:random_init}.

The largest difference was observed for the ACE 2005 Entities dataset: Using one seed value, the network achieved an $F_1$ performance of 82.5\% while using another seed value, the network achieved a performance of only 74.3\%. Even though this is a rare extreme case, the median difference between different weight initializations is still large. For example for the CoNLL 2003 NER dataset, the median difference is at 0.38\% and the 95th percentile is at 1.08\%. 

In conclusion, if the fact of different local minima is not taken care of and single performance scores are compared, there is a high chance of drawing false conclusions and either rejecting promising approaches or selecting weaker approaches due to a more or less favorable sequence of random numbers.

\section{Experimental Setup}\label{sec:experiment_setup}

In order to find LSTM-network architectures that perform robustly on different tasks, we selected five classical NLP tasks as benchmark tasks: Part-of-Speech tagging (\textit{POS}), Chunking, Named Entity Recognition (\textit{NER}), Entity Recognition (\textit{Entities}) and Event Detection (\textit{Events}).

For Part-of-Speech tagging, we use the benchmark setup described by \newcite{Toutanova_2003}. Using the full training set for POS tagging would hinder our ability to detect design choices that are consistently better than others. The error rate for this dataset is approximately 3\% \cite{Marcus_1993}, making all improvements above 97\% accuracy likely the result of chance. A 97.24\% accuracy was achieved by \newcite{Toutanova_2003}. Hence, we reduced the training set size from over 38.000 sentences to the first 500 sentences.  This decreased the accuracy to about 95\%.

For Chunking, we use the CoNLL 2000 shared task setup. For Named Entity Recognition (NER), we use the CoNLL 2003 setup. The ACE 2005 entity recognition task annotated not only named entities, but all words referring to an entity, e.g.\ the phrase \textit{U.S. president}. We use the same data split as \newcite{Li_2013}. For the Event Detection task, we use the TempEval3 Task B setup. There, the smallest extent of text, usually a single word, that expresses the occurrence of an event, is annotated.

For the POS-task, we report accuracy and for the other tasks we report the $F_1$-score.

\subsection{Model}\label{sec:lstm_model}
We use a BiLSTM-network for sequence tagging as described in \cite{Huang2015,Ma2016,Lample2016}. To be able to evaluate a large number of different network configurations, we optimized our implementation for efficiency, reducing by a factor of 6 the time required per epoch compared to \newcite{Ma2016}.

\subsection{Evaluated Parameters}\label{sec:eval_parameters}

We evaluate the following design choices and hyperparameters:\\
\textbf{Pre-trained Word Embeddings.} We evaluate the Google News embeddings (\textit{G. News})\footnote{\url{https://code.google.com/archive/p/word2vec/}}  from \newcite{word2vec}, the Bag of Words (\textit{Le. BoW}) as well as the dependency based embeddings (\textit{Le. Dep.})\footnote{\url{https://levyomer.wordpress.com/2014/04/25/dependency-based-word-embeddings/}} by \mbox{\newcite{Levy_dep_embeddings}}, three different GloVe embeddings\footnote{\url{http://nlp.stanford.edu/projects/glove/}} from \newcite{glove} trained either on Wikipedia 2014 + Gigaword 5 (\textit{GloVe1} with 100 dimensions and \textit{GloVe2} with 300 dimensions) or on Common Crawl (\textit{GloVe3}), and the \newcite{Komninos2016} embeddings (\textit{Komn}.)\footnote{\url{https://www.cs.york.ac.uk/nlp/extvec/}}. We also evaluate the approach of \mbox{\newcite{Fasttext}} (\textit{FastText}), which trains embeddings for n-grams with length 3 to 6. The embedding for a word is defined as the sum of the embeddings of the n-grams. 

\textbf{Character Representation.} We evaluate the approaches of \newcite{Ma2016} using Convolutional Neural Networks (CNN) as well as the approach of \newcite{Lample2016} using LSTM-networks to derive character-based representations.

\textbf{Optimizer.} Besides Stochastic Gradient Descent (\textit{SGD}), we evaluate \textit{Adagrad} \cite{Adagrad}, \textit{Adadelta} \cite{Adadelta}, \textit{RMSProp} \cite{RMSProp}, \textit{Adam} \cite{Adam}, and \textit{Nadam} \cite{Nadam}, an Adam variant that incorporates Nesterov momentum \cite{Nesterov:1983} as optimizers.

\textbf{Gradient Clipping and Normalization.} Two common strategies to deal with the exploding gradient problem are \textit{gradient clipping} \cite{Mikolov2012} and \textit{gradient normalization} \cite{Pascanu2013}. Gradient clipping involves clipping the gradient's components element-wise if it exceeds a defined threshold. Gradient normalization has a better theoretical justification and rescales the gradient whenever the norm goes over a threshold.

\textbf{Tagging schemes.} We evaluate the \textit{BIO} and \textit{IOBES} schemes for tagging segments. 

\textbf{Dropout.} We compare \textit{no dropout}, \textit{naive dropout}, and \textit{variational dropout} \cite{Gal2015}. Naive dropout applies a new dropout mask at every time step of the LSTM-layers. Variational dropout applies the same dropout mask for all time steps in the same sentence. Further, it applies dropout to the recurrent units. We evaluate the dropout rates $\{0.05, 0.1, 0.25, 0.5\}$.

\textbf{Classifier.} We evaluate a Softmax classifier as well as a CRF classifier as the last layer of the network.

\textbf{Number of LSTM-layers.} We evaluated \textit{1}, \textit{2}, and \textit{3} stacked BiLSTM-layers.

\textbf{Number of recurrent units.} For each LSTM-layer, we selected independently a number of recurrent units from the set $\{25, 50, 75, 100, 125\}$. 

\textbf{Mini-batch sizes.} We evaluate the mini-batch sizes 1, 8, 16, 32, and 64.

\section{Robust Model Evaluation}\label{sec:robust_evaluation}
We have shown in \refsec{sec:evaluation_networks} that re-running non-deterministic approaches multiple times and comparing score distributions is essential to draw correct conclusions. However, to truly understand the capabilities of an approach, it is interesting to test the approach with different sets of hyperparameters for the complete network.

Training and tuning a neural network can be time consuming, sometimes taking multiple days to train a single instance of a network. A priori it is hard to know which hyperparameters will yield the best performance and the selection of the parameters often makes the difference between \textit{mediocre} and \textit{state-of-the-art} performance \cite{Hutter2014}. If an approach yields good performance only for a narrow set of parameters, it might be difficult to adapt the approach to new tasks, new domains or new languages, as a large range of possible parameters must be evaluated, each time requiring a significant amount of training time. Hence it is desirable, that the approach yields stable results for a wide range of parameters.

In order to find approaches that result in high performance and are robust against the remaining parameters, we decided to randomly sample  several hundred network configurations from the set described in \refsec{sec:eval_parameters}. For each sampled configuration, we compare different options, e.g.\ different options for the last layer of the network. For example, we sampled in total 975 configurations and each configuration was trained with a Softmax classifier as well as with a CRF classifier, totaling to 1950 trained networks.

\begin{table}[h]
\centering
\begin{tabular}{|c|c|c|c|c|}
\hline
\textbf{Dataset} & \textbf{\# Configs} & \textbf{Softmax} & \textbf{CRF}   \\ \hline
POS & 111 & 18.9\% & \textbf{81.1\%} \\
$\Delta Acc.$ &  & -0.20\% &  \\  \hline
Chunking & 229 & 4.8\% & \textbf{95.2\%} \\
$\Delta F_1$ &  & -0.38\% &  \\  \hline
NER & 232 & 9.5\% & \textbf{90.5\%} \\
$\Delta F_1$ &  & -0.66\% &  \\  \hline
Entities & 210 & 13.3\% & \textbf{86.7\%} \\
$\Delta F_1$ &  & -0.84\% &  \\  \hline
Events & 202 & \textbf{61.9\%} & 38.1\% \\
$\Delta F_1$ &  &  & -0.15\% \\  \hline
\hline
Average &  & 21.7\% & \textbf{78.3\%} \\ \hline 
\end{tabular}
\caption{Percentages of configurations where Softmax or CRF classifiers demonstrated a higher test performance.}
\label{table:demo}
\end{table}

Our results are presented in  \reftable{table:demo}. The table shows that for the NER task 232 configurations were sampled randomly and for 210 of the 232 configurations (90.5\%), the CRF setup achieved a better test performance than the setup with a Softmax classifier. To measure the difference between these two options, we compute the median of the absolute differences: Let $S_i$ be the test performance ($F_1$-measure) for the Softmax setup for configuration $i$ and $C_i$ the test performance for the CRF setup. We then compute $\Delta F_1 = \text{median}(S_1 - C_1, S_2 - C_2, \dots , S_{232}-C_{232})$. For the NER task, the median difference was $\Delta F_1 = -0.66\%$, i.e.\ the setup with a Softmax classifier achieved on average an $F_1$-score of 0.66 percentage points below that of the CRF setup. 

We also evaluated the standard deviation of the $F_1$-scores to detect approaches that are less dependent on the remaining hyperparameters and the random number generator. The standard deviation $\sigma$ for the CRF-classifier is with 0.0060 significantly lower ($p < 10^{-3}$ using Brown-Forsythe test) than for the Softmax classifier with $\sigma=0.0082$.

\section{Results}\label{sec:evaluation_results}
This section highlights our main insights in the evaluation of different design choices for BiLSTM architectures. We limit the number of results we present for reasons of brevity. Detailed information can be found in \cite{Reimers_2017_Hyperparameter}.\footnote{\url{https://public.ukp.informatik.tu-darmstadt.de/reimers/Optimal_Hyperparameters_for_Deep_LSTM-Networks.pdf}} 

\subsection{Classifier}
\reftable{table:demo} shows a comparison between using a Softmax classifier as a last layer and using a CRF classifier. The BiLSTM-CRF architecture by \newcite{Huang2015} achieves a better performance on 4 out of 5 tasks. For the NER task it further achieves a 27\% lower standard deviation (statistically significant with $p < 10^{-3}$), indicating that it is less sensitive to the remaining configuration of the network.

The CRF classifier only fails for the Event Detection task. This task has nearly no dependency between tags, as often only a single token is annotated as an event trigger in a sentence. 

\begin{figure*}[t]
\centering
  \includegraphics[width=0.7\textwidth]{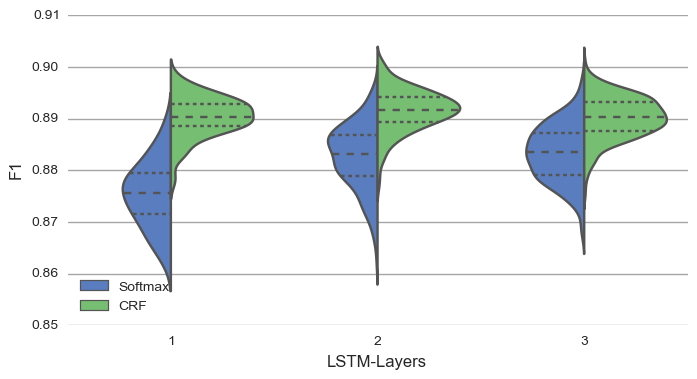}
\caption{Difference between Softmax and CRF classifier for different number of BiLSTM-layers for the CoNLL 2003 NER dataset.}
\label{fig:decoder_ner_lstm_layers}
\end{figure*}

We studied the differences between these two classifiers in terms of number of LSTM-layers. As \reffigure{fig:decoder_ner_lstm_layers} shows, a Softmax classifier profits from a deep LSTM-network with multiple stacked layers. On the other hand, if a CRF classifier is used, the effect of additional LSTM-layers is much smaller.

\subsection{Optimizer}
We evaluated six optimizers with the suggested default configuration from their respective papers. We observed that SGD is quite sensitive towards the selection of the learning rate and it failed in many instances to converge. For the optimizers SGD, Adagrad and Adadelta we observed a large standard deviation in terms of test performance, which was for the NER task at 0.1328 for SGD, 0.0139 for Adagrad, and 0.0138 for Adadelta. The optimizers RMSProp, Adam, and Nadam on the other hand produced much more stable results. Not only were the medians for these three optimizers higher, but also the standard deviation was with 0.0096, 0.0091, and 0.0092 roughly 35\% smaller in comparison to Adagrad. A large standard deviation indicates that the optimizer is sensitive to the hyperparameters as well as to the random initialization and bears the risk that the optimizer produces subpar results. 

The best result was achieved by Nadam. For 453 out of 882 configurations (51.4\%), it yielded the highest performance out of the six tested optimizers. For the NER task, it produced on average a 0.82 percentage points better performance than Adagrad.

Besides test performance, the convergence speed is important in order to reduce training time. Here, Nadam had the best convergence speed. For the NER dataset, Nadam converged on average after 9 epochs, whereas SGD required 42 epochs.

\subsection{Word Embeddings}
The pre-trained word embeddings had a large impact on the performance as shown in \reftable{table:word_embeddings}. The embeddings by \newcite{Komninos2016} resulted in the best performance for the POS, the Entities and the Events task. For the Chunking task, the dependency-based embeddings of \mbox{\newcite{Levy_dep_embeddings}} are slightly ahead of the Komninos embeddings, the significance level is at $p=0.025$. For NER, the GloVe embeddings trained on common crawl perform on par with the Komninos embeddings ($p=0.391$).
 
We observe that the underlying word embeddings have a large impact on the performance for all tasks. Well suited word embeddings are especially critical for datasets with small training sets. For the POS task we observe a median difference of 4.97\% between the Komninos embeddings and the GloVe2 embeddings.

\begin{table*}
\centering
\begin{tabular}{|c|c|c|c|c|c|c|c|c|}
\hline
\textbf{Dataset} &  \textbf{Le. Dep.} & \textbf{Le. BoW} & \textbf{GloVe1} & \textbf{GloVe2} & \textbf{GloVe3} & \textbf{Komn.} & \textbf{G. News} & \textbf{FastText}  \\ \hline
POS  & 6.5\% & 0.0\% & 0.0\% & 0.0\% & 0.0\% & \textbf{93.5\%} & 0.0\% & 0.0\% \\
$\Delta Acc.$   & -0.39\% & -2.52\% & -4.14\% & -4.97\% & -2.60\% &  & -1.95\% & -2.28\% \\  \hline
Chunking  & \textbf{60.8\%} & 0.0\% & 0.0\% & 0.0\% & 0.0\% & 37.1\% & 2.1\% & 0.0\% \\
$\Delta F_1$  &  & -0.52\% & -1.09\% & -1.50\% & -0.93\% & -0.10\% & -0.48\% & -0.75\% \\  \hline
NER  & 4.5\% & 0.0\% & 22.7\% & 0.0\% & \textbf{43.6\%} & 27.3\% & 1.8\% & 0.0\% \\
$\Delta F_1$   & -0.85\% & -1.17\% & -0.15\% & -0.73\% &  & -0.08\% & -0.75\% & -0.89\% \\  \hline
Entities & 4.2\% & 7.6\% & 0.8\% & 0.0\% & 6.7\% & \textbf{57.1\%} & 21.8\% & 1.7\% \\
$\Delta F_1$   & -0.92\% & -0.89\% & -1.50\% & -2.24\% & -0.80\% &  & -0.33\% & -1.13\% \\  \hline
Events  & 12.9\% & 4.8\% & 0.0\% & 0.0\% & 0.0\% & \textbf{71.8\%} & 9.7\% & 0.8\% \\
$\Delta F_1$   & -0.55\% & -0.78\% & -2.77\% & -3.55\% & -2.55\% &  & -0.67\% & -1.36\% \\ \hline
\hline
Average &  17.8\% & 2.5\% & 4.7\% & 0.0\% & 10.1\% & \textbf{57.4\%} & 7.1\% & 0.5\% \\ \hline   
\end{tabular}
\caption{Randomly sampled configurations were evaluated with 8 possible word embeddings. 108 configurations were sampled for POS, 97 for Chunking, 110 for NER, 119 for Entities, and 124 for Events.}
\label{table:word_embeddings}
\end{table*}

Note we only evaluated the pre-trained embeddings provided by different authors, but not the underlying algorithms to generate these embeddings. The quality of word embeddings depends on many factors, including the size, the quality, and the preprocessing of the data corpus. As the corpora are not comparable, our results do not allow concluding that one approach is superior for generating word embeddings.

\subsection{Character Representation} \label{sec:char_representation}

We evaluate the approaches of \newcite{Ma2016} using Convolutional Neural Networks (CNN) as well as the approach of \newcite{Lample2016} using LSTM-networks to derive character-based representations.

\reftable{table:char_representation} shows that character-based representations yield a statistically significant difference only for the POS, the Chunking, and the Events task. For NER and Entity Recognition, the difference to not using a character-based representation is not significant ($p > 0.01$).

The difference between the CNN approach by \newcite{Ma2016} and the LSTM approach by \newcite{Lample2016} to derive a character-based representations is statistically insignificant for all tasks. This is quite surprising, as both approaches have fundamentally different properties: The CNN approach from \newcite{Ma2016} takes only trigrams into account. It is also position independent, i.e.\ the network will not be able to distinguish between trigrams at the beginning, in the middle, or at the end of a word, which can be crucial information for some tasks. The BiLSTM approach from \newcite{Lample2016} takes all characters of the word into account. Further, it is position aware, i.e.\ it can distinguish between characters at the start and at the end of the word. Intuitively, one would think that the LSTM approach by Lample et al. would be superior.

 \begin{table}[h]
 \centering
 \begin{tabular}{|c|c|c|c|c|}
 \hline
 \textbf{Task} & \textbf{No} & \textbf{CNN} & \textbf{LSTM}     \\ \hline
POS &  4.9\% & \textbf{58.2\%} & 36.9\% \\
$\Delta Acc.$ &  -0.90\% &  & -0.05\% \\  \hline
Chunking &  13.3\% & 43.2\% & \textbf{43.6\%} \\
$\Delta F_1$   & -0.20\% & -0.00\% &  \\  \hline
NER &  27.2\% & \textbf{36.4\%} & 36.4\% \\
$\Delta F_1$ & -0.11\% &  & -0.01\% \\ \hline 
Entities & 26.8\% & 36.0\% & \textbf{37.3\%} \\
$\Delta F_1$ &   -0.07\% & 0.00\% &  \\  \hline
Events &  20.5\% & 35.6\% & \textbf{43.8\%} \\
$\Delta F_1$  & -0.44\% & -0.04\% &  \\  
\hline
Average &  18.5\% & \textbf{41.9\%} & 39.6\% \\ \hline 
 \end{tabular}
 \caption{Comparison of not using character-based representations and using CNNs \protect\cite{Ma2016} or LSTMs \protect\cite{Lample2016} to derive character-based representations. 225 configurations were sampled for POS, 241 for Chunking, 217 for NER, 228 for Entities, and 219 for Events.}
 \label{table:char_representation}
 \end{table}

\subsection{Gradient Clipping and Normalization}
For \textit{gradient clipping}  \cite{Mikolov2012} we couldn't observe any improvement for the  thresholds of 1, 3, 5, and 10 for any of the five tasks.

\textit{Gradient normalization} has a better theoretical justification \cite{Pascanu2013} and we can confirm with our experiments that it performs better. Not normalizing the gradient was the best option only for 5.6\% of the 492 evaluated configurations (under null-hypothesis we would expect 20\%). Which threshold to choose, as long as it is not too small or too large, is of lower importance. In most cases, a threshold of 1 was the best option (30.5\% of the cases).

We observed a large performance increase compared to not normalizing the gradient. The median increase was between 0.29 percentage points $F_1$-score for the Chunking task and  0.82 percentage points for the POS task.

\subsection{Dropout}
Dropout is a popular method to deal with overfitting for neural networks \cite{ Srivastava2014}. We could observe that \textit{variational dropout} \cite{Gal2015} clearly outperforms \textit{naive dropout} and not using dropout. It was the best option in 83.5\% of the 479 evaluated configurations. The median performance increase in comparison to not using dropout was between 0.31 percentage points for the POS-task and 1.98 for the Entities task. We also observed a large improvement in comparison to naive dropout between 0.19 percentage points for the POS task and 1.32 percentage points for the Entities task. Variational dropout showed the smallest standard deviation, indicating that it is less dependent on the remaining hyperparameters and the random number sequence. 

We further evaluated whether variational dropout should be applied to the output units of the LSTM-network, to the recurrent units, or to both. We observed that applying dropout to both dimensions gave in most cases (62.6\%) the best results. The median performance increase was between 0.05 percentage points and 0.82 percentage points.

\subsection{Further Evaluated Parameters}
The tagging schemes \texttt{BIO} and \texttt{IOBES} performed on par for 4 out of 5 tasks. For the Entities task, the \texttt{BIO} scheme significantly outperformed the \texttt{IOBES} scheme for 88.7\% of the tested configurations. The median difference was $\Delta F_1 = -1.01\%$. 

For the evaluated tasks, 2 stacked LSTM-layers achieved the best performance. For the POS-tagging task, 1 and 2 layers performed on par. For flat networks with a single LSTM-layer, around 150 recurrent units yielded the best performance. For networks with 2 or 3 layers, around 100 recurrent units per network yielded the best performance. However, the impact of the number of recurrent units was extremely small.

For tasks with small training sets, smaller mini-batch sizes of 1 up to 16 appears to be a good choice. For larger training sets sizes of 8 - 32 appears to be a good choice. Mini-batch sizes of 64 usually performed worst.

\section{Conclusion}

In this paper, we demonstrated that the sequence of random numbers has a \textit{statistically significant} impact on the test performance and that wrong conclusions can be made if performance scores based on single runs are compared. We demonstrated this for the two recent state-of-the-art NER systems by \newcite{Ma2016} and \newcite{Lample2016}. Based on the published performance scores, Ma and Hovy draw the conclusion of a significant improvement over the approach of Lample et al. Re-executing the provided implementations with different seed values however showed that the implementation of Lample et al.\ results in a superior score distribution generalizing better to unseen data.

Comparing score distributions reduces the risk of rejecting promising approaches or falsely accepting weaker approaches. Further it can lead to new insights on the properties of an approach. We demonstrated this for ten design choices and hyperparameters of LSTM-networks for five tasks.

By studying the standard deviation of scores, we estimated the dependence on hyperparameters and on the random seed value for different approaches. We showed that SGD, Adagrad and Adadelta have a higher dependence than RMSProp, Adam or Nadam. We have shown that variational dropout also reduces the dependence on the hyperparameters and on the random seed value. As future work, we will investigate if those methods are either less dependent on the hyperparameters or are less dependent on the random seed value, e.g.\ if they avoid converging to bad local minima.

By testing a large number of configurations, we showed that some choices consistently lead to superior performance and are less dependent on the remaining configuration of the network. Thus, there is a good chance that these configurations require less tuning when applied to new tasks or domains.

\section*{Acknowledgements}
This work has been supported by the German Research Foundation as part of the Research Training Group Adaptive Preparation of Information from Heterogeneous Sources (AIPHES) under grant No. GRK 1994/1. Calculations for this research were conducted on the Lichtenberg high performance computer of the TU Darmstadt.

\newpage

\FloatBarrier
\bibliography{references}
\bibliographystyle{emnlp_natbib}

\end{document}